\documentclass[lettersize,journal]{IEEEtran}
\usepackage{amsmath,amsfonts}
\usepackage{algorithmic}
\usepackage{algorithm}
\usepackage{array}
\usepackage[caption=false,font=normalsize,labelfont=sf,textfont=sf]{subfig}
\usepackage{textcomp}
\usepackage{stfloats}
\usepackage{url}
\usepackage{verbatim}
\usepackage{graphicx}
\usepackage{cite}
\usepackage{mathrsfs}
\usepackage{amssymb}
\usepackage{footnote}
\usepackage{multirow}
\usepackage{tabularx}
\usepackage{booktabs}
\usepackage{utfsym}
\usepackage{bm}
\usepackage{mathrsfs}
\usepackage{pifont}
\begin{document}

\title{Improving vision-language alignment with \\graph spiking hybrid Networks}

\author{Siyu~Zhang$^{\ast}$,
        Wenzhe~Liu,
         Yeming~Chen,
        Yiming~Wu,
        Heming~Zheng,
        and~Cheng~Cheng
\thanks{Siyu Zhang, Yeming Chen, and Yiming Wu are with the Department of Computer Science and Technology, Tongji University, Shanghai 201804, China (e-mail: zsyzsy@tongji.edu.cn; 2130769@tongji.edu.cn; 2410937@tongji.edu.cn).}
\thanks{Wenzhe Liu is with Department of Computer Science and Technology, Huzhou University, Huzhou 313000, China (e-mail: wenzheliu@zjhu.edu.cn).}
\thanks{Heming Zheng is with Department of Automation, Northeastern University, Shenyang 110819, China (e-mail: hemingzheng2006@163.com).}
\thanks{Cheng Cheng is with Department of Computer Science and Technology, Liaoning Normal University, Dalian 116029, China (e-mail: chengcheng@lnnu.edu.cn).}
\thanks{$\ast$ Corresponding author.}}
%\thanks{Manuscript received April 19, 2021; revised August 16, 2021.}}

\markboth{}%
{Shell \MakeLowercase{\textit{et al.}}: A Sample Article Using IEEEtran.cls for IEEE Journals}


\maketitle

\begin{abstract}
To bridge the semantic gap between vision and language (VL), it is necessary to develop a good alignment strategy, which includes handling semantic diversity, abstract representation of visual information, and generalization ability of models. Recent works use detector-based bounding boxes or patches with regular partitions to represent visual semantics. While current paradigms have made strides, they are still insufficient for fully capturing the nuanced contextual relations among various objects. This paper proposes a comprehensive visual semantic representation module, necessitating the utilization of panoptic segmentation to generate coherent fine-grained semantic features. Furthermore, we propose a novel Graph Spiking Hybrid Network (GSHN) that integrates the complementary advantages of Spiking Neural Networks (SNNs) and Graph Attention Networks (GATs) to encode visual semantic information. Intriguingly, the model not only encodes the discrete and continuous latent variables of instances but also adeptly captures both local and global contextual features, thereby significantly enhancing the richness and diversity of semantic representations. Leveraging the spatiotemporal properties inherent in SNNs, we employ contrastive learning (CL) to enhance the similarity-based representation of embeddings. This strategy alleviates the computational overhead of the model and enriches meaningful visual representations by constructing positive and negative sample pairs. We design an innovative pre-training method, Spiked Text Learning (STL), which uses text features to improve the encoding ability of discrete semantics. Experiments show that the proposed GSHN exhibits promising results on multiple VL downstream tasks. 
\end{abstract}


\section{Introduction}
\IEEEPARstart{V}{ision}-language (VL) aligned representation is one of the serious challenges within the current multimodal domain, which aims to encode image and text features and map them into a shared space, facilitating robust cross-modal semantic comprehension and interaction. VL-based multimodal interaction technology has been successfully applied in dialogue systems, autonomous driving, and assisted medical. In recent years, substantial advancements in Transformer and self-supervised learning have promoted the emergence of a large number of methods based on vision-language pre-training (VLP). VLP, as an important training paradigm, promotes cross-modal representation of large-scale image-text pairs. It has a wide range of downstream application potentials, such as Visual Question Answering (VQA) \cite{ref1}, Visual Entailment (VE) \cite{ref2}, Image-Text Retrieval (ITR) \cite{ref3}, and Natural Language Visual Reasoning (NLVR) \cite{ref4}.

The core of the VLP paradigm is modal fusion, which includes intra-modal and inter-modal fusion. The fusion mode can be categorized into two types: \textit{i)} Dual-stream model, which integrates two independent unimodal encoders to process VL information separately. \textit{ii)} Single-stream model, which aims to learn a unified representation of VL and subsequently input them into the Transformer. This model omits the architectural design of the fusion process in the dual-stream model, which is the strategy adopted by most current VLP models \cite{ref5}. Early visual embeddings relied on box-level features from predefined detectors \cite{ref6}. However, this method has some issues that cannot be ignored. First, the coarse box-level representation contains redundant information such as noise or blurred pixels (\textit{i.e.}, overlapping boundaries). Second, the interaction between boxes is prone to interference from irrelevant backgrounds, making it difficult to provide the accurate localization of objects. Third, the local features at the box level are inadequate for predicting the holistic scene of the image, thereby hindering the effective modeling of contextual semantic relations. Recently, some works \cite{ref7}, \cite{ref8} have tried to exploit grid features as a means to transcend the aforementioned limitations. Subsequently, ViLT \cite{ref9} is inspired by ViT \cite{ref10} and introduced patch features as image embeddings. Unlike grid features, image patches are directly linearly projected to improve model efficiency. Although these embedding techniques have made good progress owing to diverse tokenization strategies, they still encounter limitations in semantic understanding. On the one hand, complex image tokenization is different from language tokens that are discrete and one-dimension arranged. This inherent distribution difference between multimodalities should not be ignored. On the other hand, embedding lengthy visual sequences from high-dimensional space significantly increases the computational burden. Given this, we believe that learning good visual representation is a key factor in aligning VL semantics.
\IEEEpubidadjcol
In this article, we propose a novel visual semantic encoding module. Ideally, the salient regions and relations of different instances should be presented comprehensively and clearly to facilitate visual comprehension. To this end, we propose to adopt panoptic segmentation \cite{ref11} to capture fine-grained image features and use them as vector tokens for subsequent embedding. To model local and global contextual relations, we construct an interwoven Graph Spiking Hybrid Network (GSHN). On the one hand, GSHN possesses a significant sparsity of SNNs and naturally generates sparse attention coefficients for efficient feature aggregation. On the other hand, GSHN involves the potential composition and hierarchical relations of different networks, which is more conducive to capturing deep visual semantic information. Considering the workload of SNN, we employ a contrastive learning (CL) method to achieve the recall of nodes with similar semantics. It has the following two advantages: \textit{i)} effectively avoids the membrane potential reset problem caused by the input of a single sample. \textit{ii)} The integrated similar samples can be regarded as a “\textit{video frame}” as the input of SNN, which facilitates enhancing the capacity of SNN to capture spatiotemporal information and speeds up the computational efficiency of the model. Furthermore, we develop a Spiked Text Learning (STL) pre-training task for aligning VL semantic labels, which is adept at refining discrete semantic inputs by learning textual features. We conduct extensive experiments and demonstrate effectiveness on public datasets for multiple VL tasks, including VQA, VE, image-text retrieval, and VR. Ablation experiments further verify the rationality and effectiveness of our model design.

In summary, the main contributions of this work are as follows:
\begin{itemize}[\IEEEsetlabelwidth{Z}]
\item We propose an abstract visual-semantic representation module that refines the visual input and encodes rich and compact high-level semantic features.
\item We design a novel Graph Spiking Hybrid Network (GSHN), which studies the joint learning of SNN-based discrete semantic modules and GAT-based continuous semantic modules, aiming to enforce the intrinsic processing ability of spatiotemporal information in a robust, hierarchical, and multi-domain manner.
\item We use contrastive learning to model the semantic commonalities of similar node inputs, which helps avoid model overfitting on limited data and improves the computational efficiency of the model.
\item We design a spiked text learning (STL) pre-training task to supervise the output of SNN, thereby facilitating deep semantic mining of text features. Importantly, experiments verify the successful application of this method on multiple downstream tasks and achieve competitive results.
\end{itemize}

The rest of this paper is organized as follows. In Section II, we briefly introduce related work. In Section III, we elaborate on the proposed visual semantics module and GSHN. In Section IV, we further analyze and discuss the experimental results. Finally, Section V provides a conclusion and next plans.

\section{Related Work}
\subsection{Visual Tokenization and Embedding} 
Both vision and text involve tokenization and embedding processes. However, in contrast to the discrete, concise, and abstract language tokens, visual representations encompass more diverse and intricate information, attributable to their pixel continuity and high dimensionality. To follow text embedding, the image tokenization process is typically divided into three stages: \textit{i)} Region-based, requiring pre-trained object detectors to extract features. Notably, most VLP models employ Faster R-CNN as the visual embedding. For example, Anderson \textit{et al}. \cite{ref6}  adopted region features to train VL tasks. \textit{ii)} Grid-based, wherein some works attempt to partition the image into uniform grids and directly perform convolution operations to encode features. For instance, Jiang \textit{et al}. \cite{ref7}  used grid features to train a VQA model. Huang \textit{et al}. \cite{ref8}  proposed the Pixel-BERT method, which only randomly trains some pixels to improve computational efficiency. Later on, Huang \textit{et al}. \cite{ref12}  designed a visual dictionary to encode compact visual features. Although this method is straightforward to implement and does not require a pre-defined object detector, the deepening of the network layers and the redundant information will increase the computational complexity. \textit{iii)} Patch-based, it obtains features through the linear projection of image patches. Kim \textit{et al}. \cite{ref9}  proposed a VLP model based on the Vision Transformer (ViT). In addition, some recent large models \cite{ref13},\cite{ref14} have adopted ViT as visual embedding. Although the primary advantage of this method lies in its efficiency, the use of fixed-size patches that span across multiple visual regions may ignore the detailed features of object boundaries.

\subsection{VL-aligned representation} 
There exists an essential difference in spatial dimensions between vision and language. Achieving a good VL-aligned representation is the core goal of the VLP task. The dual-stream and single-stream modes are the primary structure designs of existing VLP methods. The dual-stream first encodes independent vision and language features. Then, the Transformer is used to align the VL representation semantic space at a high level. It has various network layers and structure configurations, which help in addressing complex VL tasks. Tan \textit{et al}. \cite{ref15} proposed LXMERT to model the relations between two modes. It is a large-scale Transformer model that mainly contains three encoders: object relations, language, and cross-modal. Yu \textit{et al}. \cite{ref16} presented a knowledge-enhanced method, named ERNIE-ViL, which endeavors to establish intricate semantic connections across vision and language from scene graphs. Li \textit{et al}. \cite{ref17} developed a novel model based on 12 datasets to explore the relations between vision and language. Single-stream directly fuses vision and language at the feature level as input to the Transformer. Gan \textit{et al}. \cite{ref18} introduced the VILLA model, which uses adversarial training to achieve VL-aligned representation learning. Huang \textit{et al}. \cite{ref12} proposed SOHO, which constructs a visual dictionary to generate visual features and achieves cross-modal comprehension by aligning language tokens. Kim \textit{et al}. \cite{ref9} proposed a vision and language Transformer (ViLT), which speeds up the training of VLP models by simplifying visual input. In addition, Li \textit{et al}. \cite{ref19} proposed SemVLP by mixing two modes, which can jointly align low-level and high-level semantics between image and text representations.

\subsection{Spiking Neural Networks} 
The Spiking neural network (SNN) \cite{ref20} is regarded as the third-generation evolution within the artificial neural network (ANN). It has attracted widespread attention due to its low power consumption and uniqueness in processing sparse data. In recent years, SNNs have been successfully applied to challenging visual tasks such as object detection, recognition, and segmentation. As researchers pursue the development of higher-performance SNNs, more and more novel methods have been proposed. Kim \textit{et al}. \cite{ref21} introduced Spiking-YOLO, the first SNN model for object detection through spike-based mechanisms. Subsequently, several works have attempted to explore deep spiking convolutional networks. Lee \textit{et al}. \cite{ref22} used spike-based backpropagation to directly train deep convolutional SNNs, which achieved comparable performance to ANNs in recognition tasks. Hu \textit{et al}. \cite{ref23} proposed a spiking version of the deep residual network (ResNet). Furthermore, they provide a compensation mechanism to alleviate the discretization errors accumulated by deep SNNs. Recently, some attempts have been made to extend SNNs to the graph domain. Zhu \textit{et al}. \cite{ref24} proposed an end-to-end SpikingGCN, which integrates graph convolution with the fidelity properties of SNN. In this framework, the original graph data is encoded as a sequence of spikes. Some works combine SNN with Transformer. For example, Zhang \textit{et al}. \cite{ref25} developed a spike transformer network (STNet) for tracking single objects, which can dynamically extract and fuse spatial and temporal information. Wang \textit{et al}. \cite{ref26} proposed the masked spiking Transformer (MST) framework, which utilizes the random spike masking (RSM) method to reduce energy consumption. Zhou \textit{et al}. \cite{ref27} first adopted pure SNN to construct a spiking visual Transformer.
%%%%%%%%%%%%%%%%%%
\section{Methods}
In this section, we initially present an overview of the proposed GSHN, encompassing key components such as the visual embedding process, continuous semantic encoder, discrete semantic encoder, and hybrid semantic transmission mechanism. Then, we perform a unified VL alignment process and introduce the STL task. The overall training procedure of GSHN is depicted in Algorithm 1.
%%%%%% fig1%%%%%%%%
\begin{figure*}[t!]
\begin{center}
\includegraphics[width=5.8in]{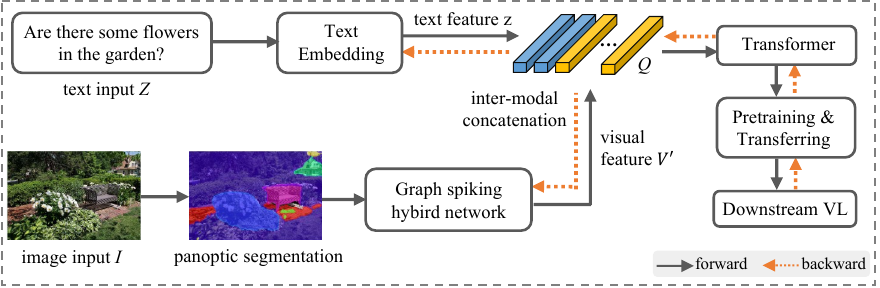}
\end{center}
   \caption{Overview of the GSHN architecture. We use panoptic segmentation to optimize the fine-grained image semantic representations. We also adopt multi-layer Transformers to apply multimodal connections on pre-training tasks, where the pre-training task follows existing ITM, MLM, and CL works.}  
\label{fig_1}
\end{figure*}

\subsection{Problem Statement}
Text tokenization can flexibly express more intricate semantic information due to the discretization and additivity in its semantic space. In light of this, we argue that it is necessary to construct a discrete image semantic space to bridge the VL semantic gap. Similar to text tokenization, we treat mask instance input as the “\textit{sentence}” instead of a rough hard mapping of the “\textit{word}”, which will reflect its intrinsic structural relations by refining the semantic interactions of each mask instance. We introduce an efficient GSHN model, the design of which is motivated by a hybrid architecture of integrating both latent continuous and discrete semantic variables. Specifically, the mask instance after segmentation optimization can be viewed as a graph node, and we first use GAT to encode graph node features to process continuous weight vectors. Then, we use the spike sequence of SNN to process discrete weight computation. Leveraging the special sparsity, energy efficiency, and robustness of SNN, we combine SNN and GAT to build a visual semantic module for aligned text. This strategy draws on the complementary advantages of hybrid to constrain the information distribution of the original space.

Modeling high-quality hybrid networks is the core goal of this work. Traditional GNNs are mainly used to develop static graphs, while SNNs have received widespread attention due to their good spatiotemporal dynamics and ability to process time-series data. Given the differences in their processing methods, it is not suitable to directly employ GNN-based workloads and evaluation metrics to verify SNNs. To this end, we need to focus on the following two aspects: \textit{i)} Developing adaptive workloads for SNN, including optimizing input, output, and corresponding pre-training tasks. \textit{ii)} Enhancing the inductive ability of the GSHN architecture, ensuring it can capture underlying patterns in VL. Fig. \ref{fig_1} presents a comprehensive overview of the proposed model. First, panoptic segmentation is performed on the original image to obtain tokenized input. Then, a dynamic scale-fused GSHN is constructed to encode continuous and discrete node feature representations. Here, we construct a selectable semantic memory unit based on the spike activation output of the SNN and extract the optimized high-level visual semantics by training the information weight ratio. Finally, we perform VL alignment to achieve an end-to-end pre-training task.

\subsection{Visual Embedding}
To represent the completeness and coherence of image scene comprehension, different from previous methods based on rectangular bounding box detection and grid features, we obtain fine-grained image representation based on panoptic segmentation. It is worth noting that panoptic segmentation extracts feature vectors by performing operations at the mask level (\textit{i.e.}, instance shape), which avoids the negative effects brought by irrelevant or overlapping regions around the objects. Specifically, given an image $I$, each instance in the image will generate a candidate segment $Mask$ (\textit{i.e.}, a set of binary masks) after the segmentation module.
\begin{equation}
\label{eq:1}  
Mask= Segment(I) 
\end{equation}
where $Mask = {mask_1,\dots, mask_L}$, $mask_i\in\{0,1\}^{H\times W}$ represents a mask of which pixels belong to an instance. $L$ is the number of masks. $H$ and $W$ are the height and width of the image, respectively. It is worth noting that the Mask contains the thing class predicted by instance segmentation and the stuff class predicted by semantic segmentation. Furthermore, the output category prediction is omitted.

Meanwhile, inspired by the Convolutional Feature Masking (CFM) method \cite{ref28}, we obtain the feature map of the image from the convolutional layer of a pre-trained CNN and resize the mask to match the size of the feature map. The mask instance feature $V$ obtained by segmentation is expressed as follows:
\begin{equation}
\label{eq:2}  
V= CNN(I)\odot resize(Mask)
\end{equation}
where $V =\{v_1,\dots, v_k\}, v_i\in\mathbb{R}^D$. $k$ represents the number of segmented regions.   represents an element-wise product operation on each channel of the feature map. Here, $V= [V^t, V^s]$, which contains thing class $V^t\{v_1^t,\dots,v^{t}k^t\},v_i^t\in\mathbb{R}^D$ and stuff class $V^s\{v_1^s,\dots,v^{s}k^s\},v_i^s\in\mathbb{R}^D$.

\subsection{Graph Spiking Hybird network}
\textbf{\textit{Continuous visual encoder:}} To preserve rich visual semantics, we propose to use a trainable GAT \cite{ref29} to model continuous semantic variables. GAT, as a variant of GNN, aims to design adaptive edge weights of the self-attention mechanism and update its embedding representation by aggregating its neighbor nodes. More specifically, given a graph $G = (X, A)$, it contains the feature vector set $X=[x_1,\dots, x_n] \in\mathbb{R}^{n\times d}$ of the graph nodes and the adjacency matrix $A=[a_{ij}]\in\mathbb{R}^{n\times n}$. Here, $a_{ij}$ represents the edge weight between nodes $i$ and $j$. $n$ is the number of nodes, and each node has $d$ dimensions. First, for each node $i$ and its neighboring node $j$, the original feature vectors $x_i$ and $x_j$ are transformed. The obtained node feature vectors are expressed as follows:
\begin{equation}
\label{eq:3}  
f_i=x_{i}W, f_j=x_{j}W
\end{equation}
where $W\in\mathbb{R}^{d\times d^{\prime}}$ represents the learnable weight matrix.

Then, the edge attention score calculation is performed and its expression is provided as follows:
\begin{equation}
\label{eq:4}  
e(f_i,f_j)=LeakyReLU([f_{i}\parallel f_j]\cdot a^T)
\end{equation}
where $LeakyReLU$ represents a nonlinear activation function, $\parallel$ represents the vector concatenation operation on the feature vectors $f_i$ and $f_j$, and $a\in\mathbb{R}^{1\times2d^{\prime}}$ represents a learnable attention parameter vector.

Next, for each node $i$, a softmax function is applied to normalize all its neighboring nodes $j\in\mathcal{N}_i$, and the attention weight $\alpha_{ij}$ is defined as:
\begin{equation}
\label{eq:5}  
\alpha_{ij}=softmax_{j}(e(f_i,f_j))=\frac{exp(e(f_i,f_j))}{\sum_{j^{\prime}\in\mathcal{N}_i}exp(e(f_i,f_{j^{\prime}}))}
\end{equation}
where $\mathcal{N}_i$ represents the set of neighboring nodes of node $i$.

Finally, the feature vector $f^{\prime}_i$ of node $i$ is updated using the normalized attention weights and the corresponding neighborhood node features.
\begin{equation}
\label{eq:6}  
f^{\prime}_i=\sigma(\sum_{j\in\mathcal{N}_i}\alpha_{ij}\cdot f_j)
\end{equation}
where $\sigma(\cdot)$ represents a nonlinear activation function.

Without box-level restrictions, we define an end-to-end learning and updating process from feature input to output. Given the input features $X$ of each batch, the formulated continuous visual encoder $\textbf{\textit V}(\cdot)$ can be expressed as:
\begin{equation}
\label{eq:7}  
F_{GAT}=\textbf{\textit V}(X; \Theta)
\end{equation}
where $F_{GAT}=[f^{\prime}_1,\dots,f^{\prime}_n]\in\mathbb{R}^{n\times d^{\prime}}$ represents the output feature. $X=[x_1,\dots, x_n]\in\mathbb{R}^{n\times d}$ represents the input graph node features. $n$ is the number of graph nodes, $d$ and $d^{\prime}$ represent the dimensions of input and output node features, respectively. $\Theta$ denotes the parameter set for training GAT.\\

\textbf{\textit{Discrete semantic encoder:}} Different from the basic information captured by continuous weight calculation in GAT, we model a discrete semantic encoder that facilitates learning inductive biases (unseen nodes or links in the graph) to generate predictions. SNNs are bio-inspired neural networks with great research potential due to their sparsity and low power consumption. SNNs utilize discrete pulses (termed spikes) rather than continuous values as carriers of information transmission and only emit a pulse when the membrane potential of the neuron reaches the threshold to transmit binary spike sequences (\textit{i.e.}, 0 or 1) and asynchronous information. In this work, SNN is incorporated into the model architecture design, aiming to better exploit spatiotemporal properties and sparsity to learn multi-network patterns with a high degree of freedom fusion. The update of membrane potential usually follows the popular Leaky Integrate-and-Fire (LIF) \cite{ref30}, which effectively simulates the spiking behavior of neurons. This method accounts for the incremental variation in input current over time, and its behavior is described as:
\begin{equation}
\label{eq:8}  
s=argmax (U_{r2},(U^{t}_{w}-U_{r1}+\Delta U_w))  
\end{equation}
where $U_{r2}$ is the signed reset potential. $U^{t}_w$ represents the membrane potential $U$ at a given time step $t$. $U_{r1}$ is the resting potential and $\Delta U_w$ represents the post-synaptic input. When the threshold $U_{r2}$ is exceeded by the membrane potential $U$, $s = 1$, otherwise $U$ is immediately reset, \textit{i.e.}, $s = 0$. The corresponding soft activated is defined as follows:
\begin{equation}
\label{eq:9}  
s=(U^{t}_{w}-U_{r1}+\Delta U_w)-U_{r2}
\end{equation}

Here, the tanh function is selected as the surrogate function, which can effectively alleviate the instability factors that occur when the pulse function performs gradient descent.

SNN needs to respond to events to process asynchronous information, which is different from $F_{GAT}$ to fix the information at each location to achieve synchronous information processing. Since $F_{GAT}$ encodes continuous variable values, while SNN accepts discrete spike inputs. To solve the feature propagation problem of SNN to deploy discrete semantic encoders, we use a popular method of probabilistic sampling to encode the input $F_{GAT}$ into a discrete spike sequence. The discrete semantic encoder $\textbf{\textit H}(\cdot)$ can be formalized as follows:
\begin{equation}
\label{eq:10}  
S=Acc^{b}_{i=1}| \textbf{\textit H} (P_{binary,i};\zeta_{i})
\end{equation}
where $S =[S_{1},\dots,S_{b}]\in\mathbb{R}^{n\times d^{\prime\prime}}$ represents the sparse spike sequence. $Acc$ means executing the accumulation operation of $P_{binary}$ from $i=1$ to $b$ into the membrane potential, $P_{binary}\in\mathbb{R}^{n\times d\times \{0,1\}\times T}$. $T$ means the time window size. $\zeta_{i}$ denotes the parameter set for training SNN. To better obtain discrete semantic features, STBP-tdBN \cite{ref31} is adopted to assist neuron training, which helps accelerate model convergence.

\subsection{Model Input and Output}
\textbf{\textit{Encoder Input:}} In this work, we use contrastive learning to capture similar graph node representations as model input, which is different from random sampling. To obtain node features with similar semantics, we need to reorder the samples with the highest matching scores in a batch calculated by ITM  \cite{ref33}, and then use the retrieve function to recall the samples. During the training process, we pre-freeze the SNN training parameters and adaptively configure epochs to ensure that only the input order is adjusted without changing the total amount of model calculation. For the continuous visual encoder, we assume that each graph is an independent sample, which is independent of other graphs. During the training of GAT, multiple graphs are set to be combined into a batch for synchronous calculation. This not only helps to improve computational efficiency but also allows the model to learn commonalities between different graphs. For the discrete semantic encoder, the input in a batch is a converted spike sequence. These spike sequences can form a visual stream with similar semantics, which as a time dimension can continuously learn events to respond to pulses, thereby realizing information transfer between graph nodes. In addition, the reorganization graph can be regarded as a hard sample sampling process, which is beneficial to the representation learning of semantics.

\textbf{\textit{Encoder Output:}} For SNN, the activation state of its neurons is accumulated, which enables SNN to remember previous input information and process historical information in the time dimension. Since SNN does not reset to the initial state after each forward propagation, we need to focus on discussing reasonable output methods rather than precise predictions. In addition, SNNs based on brain-like models have difficulty competing with GATs based on continuous computation in classification tasks, which shows that it is not appropriate to directly sum or concatenate the continuous and discrete variables of the output. Given the binary nature of spike events, SNNs prefer “\textit{yes or no}” rather than “\textit{what}” questions. To this end, we view SNNs as a selector that screens basic semantic features by activating continuous visual streams. Inspired by \cite{ref12},  we define a flexible semantic memory unit $M\in\mathbb{R}^{d^{\prime}\times d^{\prime\prime}}$ in combination with SNN, which can freely combine these basic semantics to provide more diverse features. Note that $M$ is similar to a container rather than a fixed dictionary that can only perform one query. Discrete semantic information representation can be formalized as:
\begin{equation}
\label{eq:11}  
E_{SNN}=\frac{\sum^{d^{\prime\prime}}_{j=1}M_{j}\ast S}{\sum^{n}_{i=1}S_i}
\end{equation}
where $E_{SNN}\in\mathbb{R}^{n\times d^{\prime\prime}}$, $M_j$ represents the $j$th basic semantic feature, and $S$ is the pulse feature. Here, SNN only needs to perform basic semantic feature learning within a batch. This strategy is better than continuous calculation. Since SNN is binary, when the integration time window is set to 1, the complex multiplication operation between input and weight can be omitted to achieve lower energy consumption.

%%%%%% fig2%%%%%%%%
\begin{figure*}[t!]
\begin{center}
\includegraphics[width=5.8in]{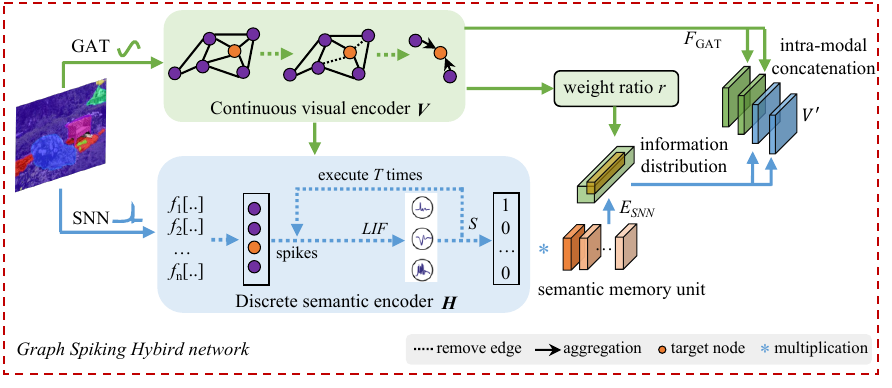}
\end{center}
   \caption{Information transmission diagram. We encode GAT-based concrete (solid green pathway) and SNN-based discrete (solid blue pathway) visual semantic modules.}  
\label{fig_2}
\end{figure*}

\textbf{\textit{Hybrid transmission:}} The proposed GSHN decouples two important structures, continuous vision, and discrete semantics. Fig. \ref{fig_2} provides the detailed framework of GSHN. When the model performs the training within a batch, the continuous output of GAT encoding needs to calculate its total information at once. Unlike SNN with spatiotemporal properties, it needs to accumulate input activation properties in the time dimension to reduce the time step of each graph spike sampling, which is more advantageous than the previous reset process. However, it is crucial to effectively fuse the output differences between the two. To this end, we design an information transfer unit with intermediate representation, which not only integrates the heterogeneous properties of GAT and SNN but also realizes information optimization with dynamic proportional fusion to bridge the gap between the two. More specifically, we first need to build a squeeze module for the continuous output $F_{GAT}$. Then, we perform a fully connected mapping to obtain the information weight ratio $r$. Next, we perform the excitation calculation on the $E_{SNN}$ with the assigned weight $r$ within each batch, thereby obtaining the final discrete semantic representation. Finally, an addition operation is performed on the continuous semantic $F_{GAT}$ and the discrete semantic $E_{SNN}$ to transmit the mixed information flow and input it into the pre-training task as the final image feature vector. The output of the GSHN can be formalized as:
\begin{equation}
\label{eq:12}  
r=\sigma\cdot\phi(F_{GAT})
\end{equation}
\begin{equation}
\label{eq:13}  
V^{\prime}=r\ast E_{SNN}+F_{GAT}
\end{equation}
where $\sigma(\cdot)$ and $\phi(\cdot)$ denote the sigmoid and fully connected (FC) functions, respectively.
%%%%%%%%%%%%%algorithm1
\begin{algorithm}[H]
	\renewcommand{\algorithmicrequire}{\textbf{Output: }Image-text representation $Q$.}{\textbf{Input: }Image $I$, Text $Z$.}
\caption{Training steps of our GSHN framework}
\label{alg:1}
\begin{algorithmic}[1]
\REQUIRE 
\STATE Segment image $I$ to extract instance $Mask$ features $V\in\mathbb{R}^D$. 
\STATE Encode GAT to obtain continuous vectors $F_{GAT}\in\mathbb{R}^{100\times 768}$.
\STATE Encode SNN to obtain discrete vectors $S\in\mathbb{R}^{100\times 3000}$, with binary values \{0,1\}.
\STATE Initialize semantic memory units $M\in\mathbb{R}^{768\times 3000}$.
\STATE Compute discrete semantic output $E_{SNN}\in\mathbb{R}^{100\times 768}$ via multiplication, acting as a “select-and-combine” manner for $M$.
\STATE Construct a squeeze module for $F_{GAT}$ and calculate the information weight ratio $r$.
\STATE Perform excitation calculation to redistribution information to obtain the final discrete semantics. 
\STATE Combine $F_{GAT}$ and the final discrete semantics via summation to obtain the image feature $V^{\prime}$.
\STATE Take $V^{\prime}$ and z (text features) as input for the multimodal alignment task.
\FOR {epoch from 1 to 40}
        \IF {epoch $\le$10}
        \STATE Freeze SNNs
        \STATE Perform pre-training tasks: ITM, MLM, CL   
        \ELSE  
        \STATE Unfreeze SNNs
        \STATE Train STL task  
        \ENDIF      
        \ENDFOR    
\RETURN  $Q$.
	\end{algorithmic}  
\end{algorithm}

\subsection{Vision-language alignment}
For the pre-training process, we implement a medium-sized multi-layer Transformer to model the joint representation of VL features, which aims to maximize the matching of the interactive features of abstract visual representation and high-level language representation, narrowing the semantic gap in the feature space. For language embedding, we adopt the BERT \cite{ref32} tokenizer to tokenize text prompts. For image embedding, we propose GSHN based on the text processing method, which maps the segmented mask instances into fine-grained “\textit{sentences}” instead of simple “\textit{words}”. We record the position of the visual output and encode it in the same way as BERT, thus obtaining VL-aligned token masks and token embeddings. We perform multiple pre-training tasks, including the existing MLM, ITM, and CL, evaluating the usefulness of GSHN. In addition, when masking a certain mask feature region, the SNN triggers neurons to produce a multi-class label, which helps our proposed STL learning. Note that we only use the text features encoded by the Transformer for prediction, aiming to better align the sparse visual and textual semantics.

We design the STL pre-training task, which performs supervised learning on the discrete output of the SNN and corrects its output. Essentially, the input of the STL task is the partial signal of the SNN and the text vector features, which are used to predict the SNN signal corresponding to the masked partial mask instance region. This process is similar to MLM, which assists images as candidates through text semantics. The specific masking rules are as follows:
\begin{equation}
\label{eq:14}  
SNN_{prob=0.05}=STL(S, others, W)
\end{equation}
where $S$ represents the output spike sequence. $W$ represents the input word, which is defined as $W=\{W_1,\dots, W_L\}$. $L$ represents the length of the text. Here, considering that the dimension of the semantic memory unit is 3000, we set a 5\% probability to mask the output of SNN.

For multi-label classification tasks, we regard it as a binary vector $Y$, where each element corresponds to a category. Once a dimension is activated, it is classified under the pertinent category, with the element at its corresponding position being allocated a value of $Y=1$. Conversely, the assigned value is $Y=0$. To assess the difference between the true label and the predicted result $Y$, we use focal loss to reduce the weight of easy-to-classify samples, so that the model pays more attention to hard samples. The calculation expression of the loss function is given by:
\begin{equation}
\label{eq:15}  
\mathcal{L}_{STL}=-\frac{1}{N}\sum\limits^{N}\sum\limits^{W}y_{i,j}log(\hat{y}_{i,j})+(1-y_{i,j})log(1-\hat{y}_{i,j})
\end{equation}

\section{Experiments}
\subsection{Implementation Details}
For the text backbone, we follow BERT to tokenize the text using the WordPiece tokenizer. We use the AdamW optimizer with an initial learning rate of $1\times10^{-4}$ and a weight decay of $1\times10^{-2}$ for optimization. The number of training iterations is 1000 and the warm-up ratio is set to $1/3$. For the visual backbone, we set the image input size to $640\times640$ to maintain a fair comparison. We use ImageNet \cite{ref35} to initialize the underlying encoder, which uses ResNet-101 as the visual embedding backbone architecture to extract image features. We use DeepLab \cite{ref36} and Mask R-CNN \cite{ref37} for semantic segmentation and instance segmentation respectively, and merge their outputs to perform mask instance feature extraction jointly. Note that the output features determine its size. The non-maximum suppression (NMS) threshold for Mask R-CNN is set to 0.5, and its confidence threshold is set to 0.6. Here, we choose SGD as the optimizer, with a learning rate of $1\times10^{-2}$ and a weight decay of $5\times10^{-4}$. For SNN, we set the momentum of SGD to 0.9 and the initial learning rate to $1\times10^{-2}$. The semantic memory capacity is set to 3000 and the dimension of each word is 768. The pre-trained Transformer architecture is composed of 12 stacked Transformer encoder layers, each of which contains two sub-layers, namely the multi-head self-attention layer and the feed-forward network, where the number of heads is 8. Each sub-layer is a residual connection structure followed by layer normalization. In addition, we adopt ITM, MLM, and CL to implement preprocessing, where MLM is applied to positive image-text pairs. To learn continuous visual features, we freeze the SNN parameter part for the first 10 epochs and execute the Transformer. Subsequently, we use CL to process batches of graph nodes to extract similar semantic information and train GSHN using STL. Our experiments are conducted on a server equipped with 8 NVIDIA A100 GPUs, and the model executes 40 epochs to converge during pre-training.
%%%%%%%%%Table1%%%%%%%%
\begin{table*}[!t]
        \centering
	\caption{Datasets statistics on various VL tasks. Notation “$\sharp$” denotes Karpathy split.}
	\label{table_1}
	\setlength{\tabcolsep}{3pt} %表格线加粗
	\renewcommand\arraystretch{1.5} %增加表格行距
	\begin{tabular}{|m{2cm}<{\centering}|m{1.8cm}<{\centering}|m{1.8cm}<{\centering}|m{1.8cm}<{\centering}|m{1.8cm}<{\centering}|m{1.8cm}<{\centering}|m{1.8cm}<{\centering}|m{1.8cm}<{\centering}|}
	%保持水平\垂直居中
		\hline 
		Task &\multicolumn{3}{c|}{VQA} & VE & NLVR &\multicolumn{2}{c|}{ITR}\\ \hline
		Dataset & VQAv2 & VQA-CP v2 & GQA & SNLI-VE & NLVR$^2$ & Flickr30K\cite{ref40} & MSCOCO \\ \hline 
		Train Split & train+val & train & train & train & train & train & train+restval$^{\sharp}$  \\ \hline
		Test Split & test-dev/test-std & test-std & test-std & val/test & dev/test-P & test$^\sharp$ & test$^\sharp$   \\ \hline
		Metric &\multicolumn{3}{c|}{VQA-score \cite{ref41}} & \multicolumn{2}{c|}{Top-1 Accuracy} & \multicolumn{2}{c|}{Recall@1, 5, 10}  \\ \hline
	\end{tabular}
%\vspace*{-20pt}
\end{table*}
%%%%%%%%%%%%%%%%%%%%%
%%%%%%%%%Table2%%%%%%%%
\begin{table*}[!t]
        \centering
	\caption{The impact of information weight ratio on the VQAv2 dataset.}
	\label{table_2}
	\setlength{\tabcolsep}{3pt} %表格线加粗
	\renewcommand\arraystretch{1.5} %增加表格行距
	\begin{tabular}{|m{1.8cm}<{\centering}|m{1.5cm}<{\centering}|m{1.5cm}<{\centering}|m{1.5cm}<{\centering}|m{1.5cm}<{\centering}|m{1.5cm}<{\centering}|m{1.5cm}<{\centering}|m{1.5cm}<{\centering}|m{1.5cm}<{\centering}|}
	%保持水平\垂直居中
		\hline 
		\multirow{2}{*}{Model} &\multicolumn{4}{c|}{Test-dev (\%)} &\multicolumn{4}{c|}{Test-std (\%)}\\ \cline{2-9}
		& Yes/No  &Number & Other  &Overall & Yes/No  &Number & Other  &Overall \\ \hline
		without $r$ & 87.63  & 56.03 & 61.08  &71.16 & 87.62 & 56.05 & 61.98 & 71.33 \\ \hline
		with $r$ & 89.91  &56.32 & 63.44  &73.39 & 89.99 & 57.09 & 66.97 & 73.56 \\ \hline
\end{tabular}
\end{table*}
%%%%%%%%%%%%%%%%%%%%%fig3
\begin{figure*}[t!]
\begin{center}
\includegraphics[width=2.6in]{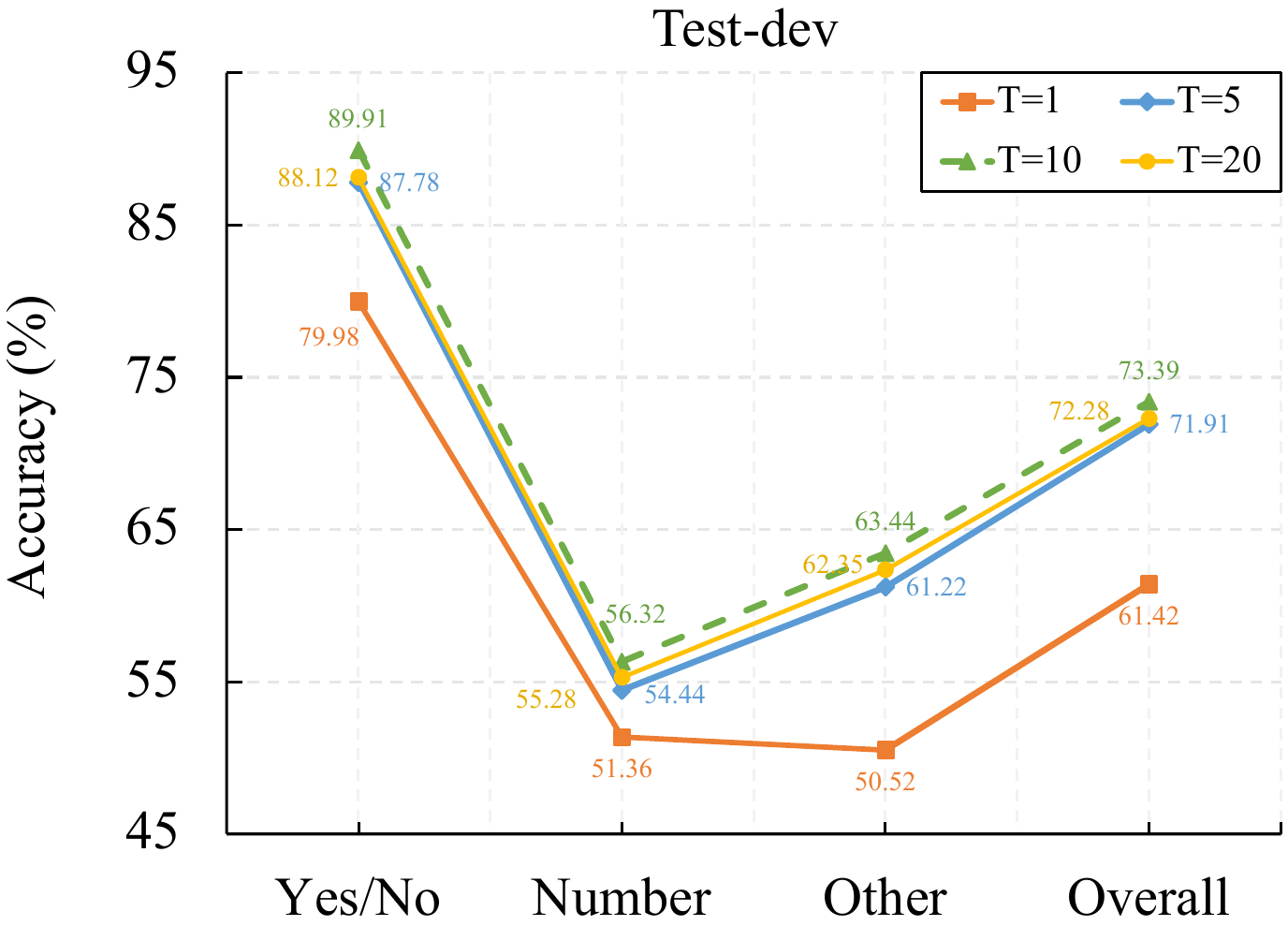}~~~~~~~~~~
\includegraphics[width=2.6in]{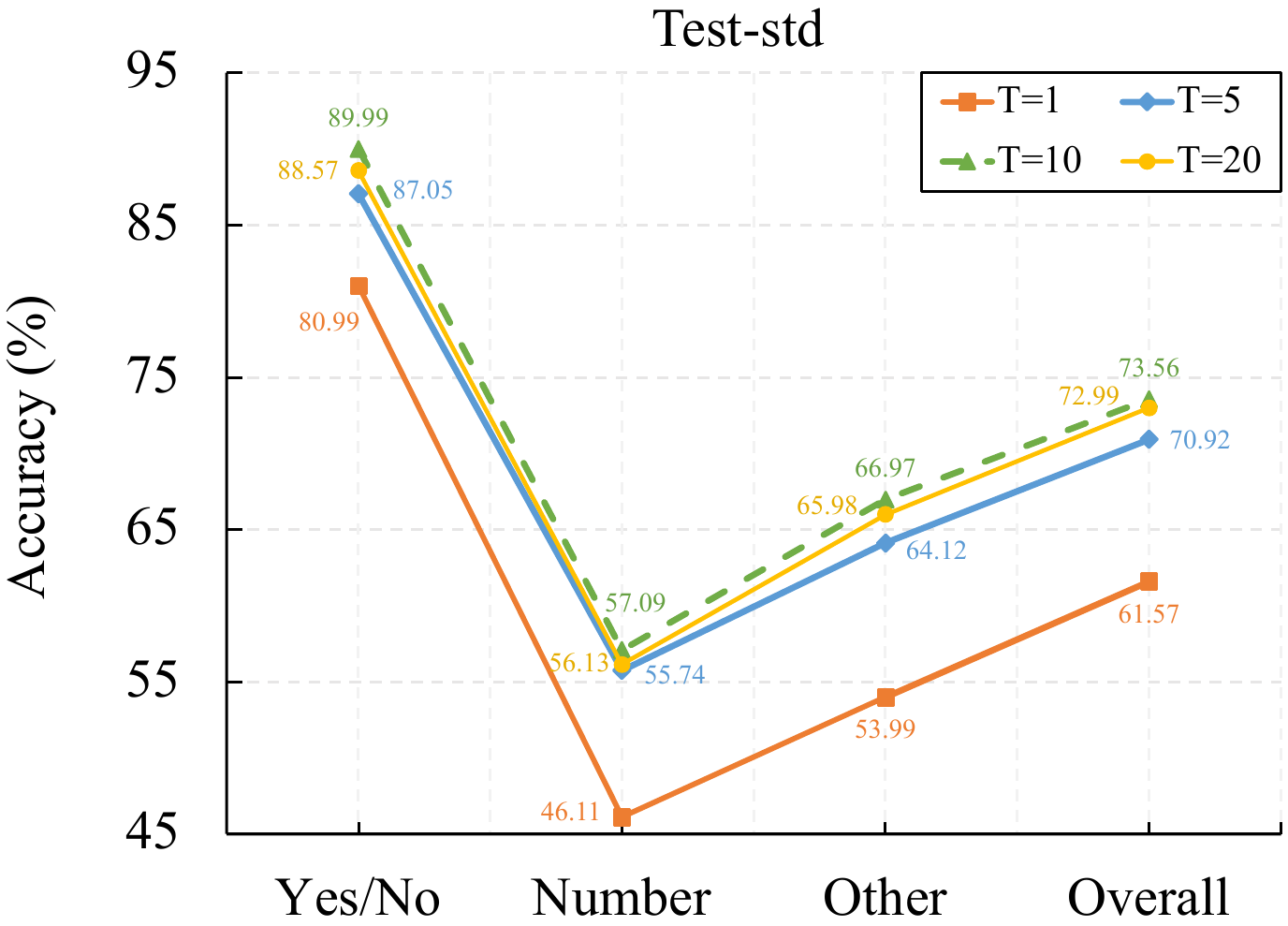}
\end{center}
   \caption{Performance testing for time window sizes on the VQAv2 dataset.}
\label{fig_3}
\end{figure*}

\subsection{Pre-training Datasets}
To demonstrate the effectiveness and generalization ability of our method, several popular datasets are used to facilitate VL pre-training for multiple downstream applications, including VQA, VE\footnote{https://github.com/necla-ml/SNLI-VE}, NLVR\footnote{https://lil.nlp.cornell.edu/nlvr/}, and ITR. The distribution of downstream datasets is summarized in Table \ref{table_1}. We pre-train on VG \cite{ref38} and MSCOCO \cite{ref39}, which are typical in-domain datasets commonly used in many VL tasks. It is worth noting that we only use part of the VG and MSCOCO datasets for training to effectively avoid the data leakage problem.
%%%%%%%%%Table3%%%%%%%%
\begin{table}[!t]
        \centering
	\caption{Performance testing of different detection methods on various VL tasks.}
	\label{table_3}
	\setlength{\tabcolsep}{3pt} %表格线加粗
	\renewcommand\arraystretch{1.5} %增加表格行距
	\begin{tabular}{|m{1.3cm}<{\centering}|m{1.2cm}<{\centering}|m{1.2cm}<{\centering}|m{1.2cm}<{\centering}|m{1.2cm}<{\centering}|m{1.2cm}<{\centering}|m{1.2cm}<{\centering}|m{1.2cm}<{\centering}|}
	%保持水平\垂直居中
		\hline 
		\multirow{2}{*}{Embedding} & VQAv2  & SNLI-VE & NLVR$^2$ &\multicolumn{2}{c|}{MSCOCO (1K) Test}\\ \cline{2-6}
		& test-dev  & test & dev  &IR & TR \\ \hline
		Grid & 69.36  & 86.03 & 80.89  &70.20 & 84.11 \\ \hline
		Bounding box & 71.72  & 87.54 & 82.46  &72.26 & 86.08 \\ \hline
	 Panoptic & 73.39  &89.22 & 84.57  &74.40 & 88.12 \\ \hline
	\end{tabular}
%\vspace*{-15pt}
\end{table}

%%%%%%%%%Table4%%%%%%%%
\begin{table}[!t]
        \centering
	\caption{The impact of different batch sizes on VL tasks.}
	\label{table_4}
	\setlength{\tabcolsep}{3pt} %表格线加粗
	\renewcommand\arraystretch{1.5} %增加表格行距
	\begin{tabular}{|m{1.3cm}<{\centering}|m{1.2cm}<{\centering}|m{1.2cm}<{\centering}|m{1.2cm}<{\centering}|m{1.2cm}<{\centering}|m{1.2cm}<{\centering}|m{1.2cm}<{\centering}|m{1.2cm}<{\centering}|}
	%保持水平\垂直居中
		\hline 
		\multirow{2}{*}{Batch size} & VQAv2  & SNLI-VE & NLVR$^2$ &\multicolumn{2}{c|}{MSCOCO (1K) Test}\\ \cline{2-6}
		& test-dev  & test & dev  &IR & TR \\ \hline
		4 & 72.92  & 86.65 & 80.62  &73.54 & 86.43 \\ \hline
		8 & 73.39  & 86.75 & 82.91  &73.93 & 87.59 \\ \hline
		16 & 73.39  &89.22 & 84.57  &74.40 & 88.12 \\ \hline
		32 & 72.03  & 87.03 & 82.74  &74.57 & 88.30 \\ \hline
		64 & 71.33  & 86.65 & 82.72 &74.71 & 88.30 \\ \hline
	\end{tabular}
%\vspace*{-10pt}
\end{table}
%%%%%%%%%Table5%%%%%%%%
\begin{table*}[!t]
        \centering
	\caption{Performance testing of four different state models.}
	\label{table_5}
	\setlength{\tabcolsep}{3pt} %表格线加粗
	\renewcommand\arraystretch{1.5} %增加表格行距
	\begin{tabular}{|m{1.8cm}<{\centering}|m{1.6cm}<{\centering}|m{1.6cm}<{\centering}|m{1.6cm}<{\centering}|m{1.6cm}<{\centering}|m{1.6cm}<{\centering}|m{1.6cm}<{\centering}|}
	%保持水平\垂直居中
		\hline 
		\multirow{2}{*}{Model} &\multirow{2}{*}{Parameter $r$} & VQAv2 & SNLI-VE & NLVR$^2$ &\multicolumn{2}{c|}{MSCOCO (1K) Test}\\ \cline{3-7}
		 &  & test-dev & test  &dev  &IR   & TR  \\ \hline
		 GAT &0  &61.08 &82.02  &80.53  &63.28   &85.00  \\ \hline
		 GAT+SNN &1  &69.96 &83.17  &81.88  &64.59   &86.20  \\ \hline
		 SNN &-  &55.51 &71.72  &70.43  &51.97   &61.07  \\ \hline
		 GSHN &trainable &73.39 &89.22  &84.57  &74.40   &88.12 \\ \hline
\end{tabular}
\end{table*}
%%%%%%%%%Table6%%%%%%%%
\begin{table*}[!t]
        \centering
	\caption{Performance testing of different pre-training tasks on the VQAv2 dataset.}
	\label{table_6}
	\setlength{\tabcolsep}{3pt} %表格线加粗
	\renewcommand\arraystretch{1.5} %增加表格行距
	\begin{tabular}{|m{2.3cm}<{\centering}|m{1.5cm}<{\centering}|m{1.5cm}<{\centering}|m{1.5cm}<{\centering}|m{1.5cm}<{\centering}|m{1.5cm}<{\centering}|m{1.5cm}<{\centering}|m{1.5cm}<{\centering}|m{1.5cm}<{\centering}|}	%保持水平\垂直居中
		\hline 
		\multirow{2}{*}{Model} &\multicolumn{4}{c|}{Test-dev (\%)} &\multicolumn{4}{c|}{Test-std (\%)}\\ \cline{2-9}
		& Yes/No  &Number & Other  &Overall & Yes/No  &Number & Other  &Overall \\ \hline
		baseline & 76.64  & 47.03& 50.02  &59.96 & 76.62 & 46.25 & 50.99 & 61.30 \\ \hline
		baseline+CL & 89.02  & 55.29 & 61.17  &72.30 &88.84  &56.02  & 64.83 & 71.91 \\ \hline
		baseline+CL+STL & 89.91  &56.32 & 63.44  &73.39 & 89.99 & 57.09 & 66.97 & 73.56 \\ \hline
\end{tabular}
\end{table*}
%%%%%% fig4%%%%%%%%
\begin{figure}[t!]
\begin{center}
\includegraphics[width=2.8in]{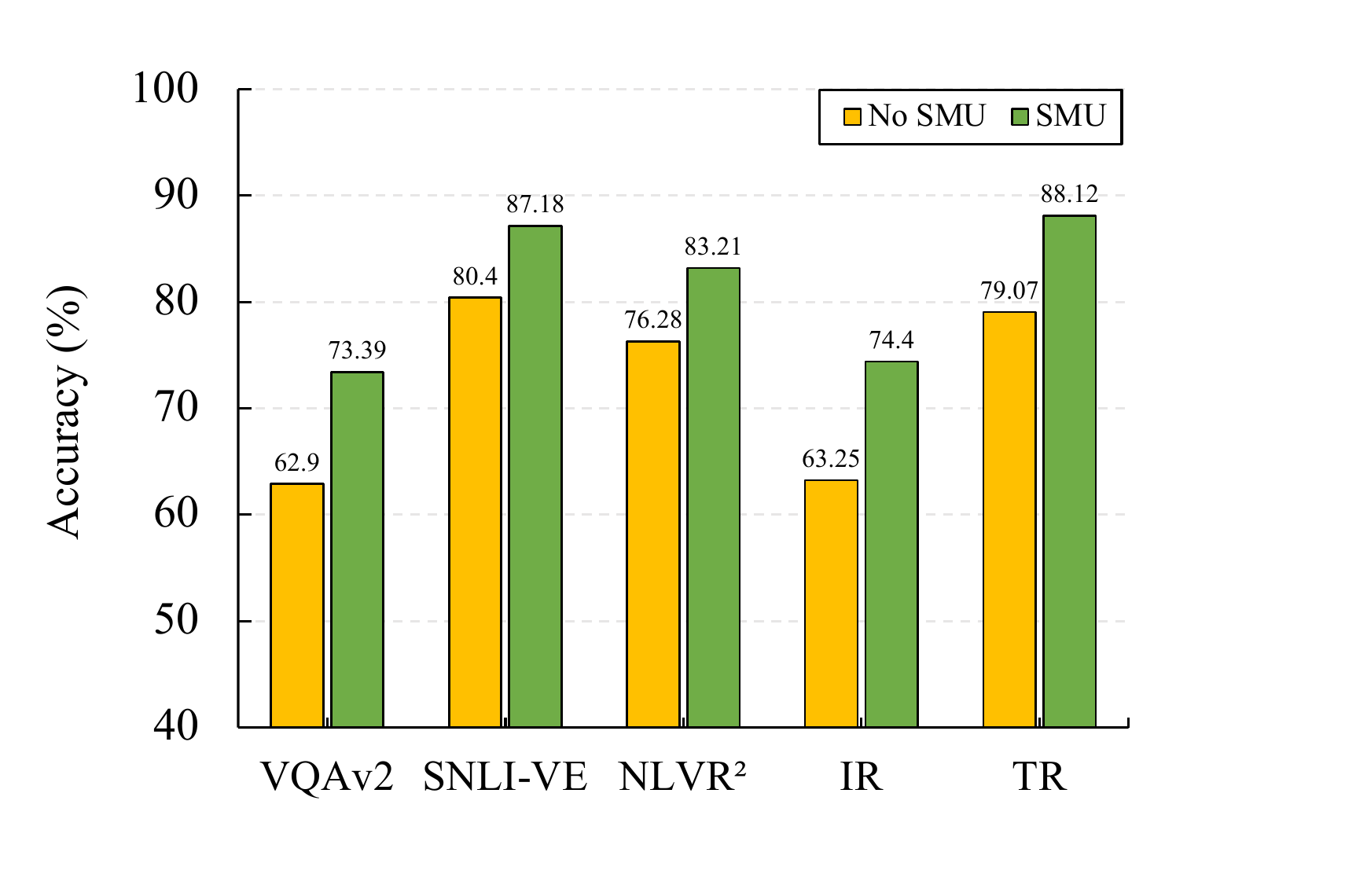}
\end{center}
   \caption{Performance comparison of the proposed GSHN with and without semantic memory unit (SMU).}
\label{fig_4}
\end{figure}

The VQA task aims to infer the correct answer by comprehending the natural language questions related to the image. It includes the currently popular VQAv2\footnote{https://visualqa.org/download.html}, VQA-CP v2\footnote{https://computing.ece.vt.edu/$\sim$aish/vqacp/}, and GQA\footnote{https://cs.stanford.edu/people/dorarad/gqa/} datasets. VQA-CP v2 is a restructured version of VQAv2, which changes the distribution of answers for each question type between the training set and the test set. They define questions into three types, namely “\textit{Yes/No}”, “\textit{Number}” and “\textit{Other}”. Compared with other datasets, GQA requires less language bias and multi-step reasoning, which is more helpful for evaluating visual reasoning ability. VE can be regarded as a three-classification problem, which is used to predict the relations between images and text (\textit{i.e.}, implicit, neutral, or contradictory). The SNLI-VE dataset is used to train the VE task. NLVR is a complex visual reasoning task that challenges the model’s ability to judge whether a sentence truly describes a visual scene, covering object collection, comparison, and comprehension of spatial relations. We use the NLVR$^2$ dataset. ITR retrieves relevant samples from another modality given an expression in one modality. It usually includes two subtasks: image-to-text retrieval (TR) and text-to-image retrieval (IR). Here, we choose the Flickr30K and MSCOCO datasets for testing.

\subsection{Ablation Study and Analysis}
We conduct ablation experiments on different downstream VL tasks to evaluate the contribution of each module in GSHN, where ResNet-101 and 3-layer Transformers are used to save computational resources.

%%%%%%%%%Table7%%%%%%%%
\begin{table*}[!t]
        \centering
	\caption{The impact of pre-training speed on the performance of different combination models.}
	\label{table_7}
	\setlength{\tabcolsep}{3pt} %表格线加粗
	\renewcommand\arraystretch{1.5} %增加表格行距
	\begin{tabular}{|m{1.9cm}<{\centering}|m{1.8cm}<{\centering}|m{1.6cm}<{\centering}|m{1.6cm}<{\centering}|m{1.6cm}<{\centering}|m{1.6cm}<{\centering}|m{1.6cm}<{\centering}|m{1.6cm}<{\centering}|}
	%保持水平\垂直居中
		\hline 
		\multirow{2}{*}{Model} &\multirow{2}{*}{Parameter (M)} &\multirow{2}{*}{Time (h)}& VQAv2 & SNLI-VE & NLVR$^2$ &\multicolumn{2}{c|}{MSCOCO (1K) Test}\\ \cline{4-8}
		 &  &  &  test-dev & test  &dev  &IR   & TR  \\ \hline
		 GAT &52.01  &54.19 &66.93  &82.12  &80.27   &69.09 &81.43  \\ \hline
		 SNN &52.32  &57.21 &57.03  &71.03  &60.99   &61.02 &62.83  \\ \hline 
		 GAT+GAT &104.02  &82.13 &69.98 &84.03 &77.32  &71.97  &83.09 \\ \hline
		 SNN+SNN &104.64  &97.42 &57.40 &72.92 &61.51  &61.90  &63.30 \\ \hline
		 GSHN &104.33 &86.17 &73.39 &89.22  &84.57  &74.40   &88.12 \\ \hline
\end{tabular}
\end{table*}
%%%%%%%%%Table8%%%%%%%%
\begin{table}[!t]
        \centering
	\caption{Comparison results of various models on the VQA v2 \\Test-dev and Test-std datasets.}
	\label{table_8}
	\setlength{\tabcolsep}{3pt} %表格线加粗
	\renewcommand\arraystretch{1.5} %增加表格行距
	\begin{tabular}{|m{1.6cm}<{\centering}|m{1.6cm}<{\centering}|m{1.4cm}<{\centering}|m{1.5cm}<{\centering}|m{1.4cm}<{\centering}|}
	%保持水平\垂直居中
		\hline 
		Model &Embedding & Transformer  & Test-dev (\%) & Test-std (\%) \\ \hline
		DAQC \cite{ref42} & &\ding{55}  &64.51&- \\ \cline{1-1} \cline{3-5}
		MLVQA \cite{ref43} &\multirow{5}{*}{Bounding box} &\ding{51} &70.30 &70.57\\ \cline{1-1} \cline{3-5}
		UNITER$_{base}$ \cite{ref5} & &\ding{51} &72.70 &72.91\\ \cline{1-1} \cline{3-5}
		BGN \cite{ref44} & &\ding{55} &72.39 &72.56\\ \cline{1-1} \cline{3-5}
		OSCAR$_{base}$ \cite{ref45} & &\ding{51} &73.16 &73.44\\ \hline
                 Pixel-BERT \cite{ref8} &\multirow{2}{*}{Grid} &\ding{55}  &71.35&71.42 \\ \cline{1-1} \cline{3-5}
		SOHO \cite{ref12} & &\ding{55}  &73.25 &73.47 \\ \hline
		ViLT \cite{ref9} &\multirow{2}{*}{Patch} &\ding{55}  &71.26 &- \\ \cline{1-1} \cline{3-5}
		PTP-ViLT \cite{ref46} & &\ding{55}  &72.13 &73.36 \\ \hline
		LOIS \cite{ref47} &\multirow{2}{*}{Instance} &\ding{51}  &72.78 &73.02 \\ \cline{1-1} \cline{3-5}
		GSHN(Ours) & &\ding{55}  &73.39 &73.56 \\ \hline				
\end{tabular}
\end{table}
%%%%%%%%%Table9%%%%%%%%
\begin{table}[!t]
        \centering
	\caption{Performance comparison of various models on the \\VQA-CP v2 dataset.}
	\label{table_9}
	\setlength{\tabcolsep}{3pt} %表格线加粗
	\renewcommand\arraystretch{1.5} %增加表格行距
	\begin{tabular}{|m{1.9cm}<{\centering}|m{1.4cm}<{\centering}|m{1.4cm}<{\centering}|m{1.4cm}<{\centering}|m{1.4cm}<{\centering}|}
	%保持水平\垂直居中
		\hline 
		Model &Yes/No & Number  & Other & Overall \\ \hline
		CAM \cite{ref48}  & 43.29 &12.31  &45.41 & 39.75 \\ \hline
		MRA-Net \cite{ref49} & 44.53 &13.05  &45.83 & 40.45 \\ \hline
		MLVQA \cite{ref43} & 40.84 &13.24  &48.83 & 41.08 \\ \hline
		LOIS \cite{ref47} & 51.82 &14.24  &41.39 & 43.09 \\ \hline
		UpDn+LPF \cite{ref50} & 88.61 &23.78  &46.57 & 55.34 \\ \hline
		GGE-DQ$_{tog}$ \cite{ref51} & 87.04 &27.75  &49.59 & 57.32 \\ \hline
		CSS \cite{ref52} & 84.37 &49.42  &48.21 & 58.95 \\ \hline		
		GSHN (Ours) &89.95  &38.75 &50.74 &58.97\\ \hline			
\end{tabular}
\end{table}
%%%%%%%%%Table10%%%%%%%%
\begin{table}[!t]
        \centering
	\caption{Comparison with other methods on the GQA dataset.}
	\label{table_10}
	\setlength{\tabcolsep}{3pt} %表格线加粗
	\renewcommand\arraystretch{1.5} %增加表格行距
	\begin{tabular}{|m{2.5cm}<{\centering}|m{1.8cm}<{\centering}|m{1.8cm}<{\centering}|}
	%保持水平\垂直居中
		\hline 
		Model &Test-dev (\%) & Test-std (\%)   \\ \hline
		LXMERT  \cite{ref15}  & 60.00 &60.33   \\ \hline
		Lyrics  \cite{ref53} & 62.40 &-   \\ \hline
		NSM  \cite{ref54} &62.95  &63.17   \\ \hline
		MDETR$_{ENB5}$ \cite{ref55} &62.95  &62.45  \\ \hline
		CuMo$_{7B}$  \cite{ref56} & 64.90 &-   \\ \hline
		VinVL$_{base}$ \cite{ref57} &65.05  &64.65   \\ \hline
		GSHN (Ours) &65.09  &64.45 \\ \hline			
\end{tabular}
\end{table}

\textbf{\textit{Information weight ratio.}} To analyze the impact of the Information weight ratio $r$, we simulated two states, \textit{i.e.}, “\textit{with r}” and “\textit{without r}”. As shown in Table \ref{table_2}, the model performance of “\textit{without r}” is lower than that of “\textit{with r}”. For the model “\textit{without r}”, the discrete semantic inputs within a batch are the same. This shows that our discrete semantic encoder focuses on the semantic content itself, but ignores the “\textit{Number}” issue. In contrast, the model “\textit{with r}” uses fewer parameters to complete the intra-modal semantic interaction, which is suitable for the design of question types “\textit{Other}” and “\textit{Yes/No}”. The weight ratio set in this paper redistributes information on discrete semantics with commonality, thereby capturing subtle differences between similar nodes.

\textbf{\textit{Embedding manner.}} Table \ref{table_3} explores the impact of different visual detection methods on model performance, covering bounding box detection, grid detection, and the panoramic segmentation embedding strategy proposed in our study. Experimental results show that panoramic segmentation has significant advantages in various VL tasks. Since this method can effectively represent the fine-grained semantics of image content, it has positively improved model performance.

\textbf{\textit{Hybrid transmission.}} We use contrastive learning to model similar node inputs, and the batch size hyperparameter setting is crucial for model training. From Table \ref{table_4}, tests on the VQAv2, SNLI-VE, and NLVR$^2$ datasets show that setting the batch size too large will reduce the robustness of the model. Since a large hyperparameter will recall too many similar nodes, it is easy to generate a large number of noisy semantics. In addition, a large hyperparameter setting will increase the memory burden of the model. On the contrary, a small batch size setting will recall a small number of similar nodes, which will lead to a waste of event-driven SNN training efficiency and is not conducive to discrete semantic inductive bias learning. For the MSCOCO dataset, a larger batch size makes it easier to obtain comparative hard samples, which is beneficial to feature retrieval. Therefore, batch size=16 is our optimal setting in this work.

\textbf{\textit{Impact of parameter r.}} As shown in Table \ref{table_5}, we compare four different state models on the downstream VL dataset. “\textit{SNN}” achieves weaker performance because it only activated discrete semantics in semantic memory units and had difficulty handling more complex VL tasks. Compared to the fixed “\textit{GAT}” and “\textit{GAT+SNN}” states, “\textit{GSHN}” adopts a variable parameter $r$ to re-match the features, and it can mine the maximum proximity of visual semantics according to the calculated weights.

\textbf{\textit{Impact of STL pre-training task.}} To further investigate the effectiveness of STL design, we train three methods, which include “\textit{baseline}”, “\textit{baseline+CL}”, and “\textit{baseline+CL+STL}”. From Table \ref{table_6}, we can see that the “\textit{baseline+CL+STL}” obviously boosts the model's performance. Specifically, the “\textit{baseline+CL+STL}” accuracy is improved by 13.43\% (from 59.96\% to 73.39\%) and 12.26\% (from 61.30\% to 73.56\%), respectively compared with the “\textit{baseline}” method. In contrast, the “\textit{baseline}” lacks common learning of features within a batch, which makes the activation output of the event-oriented SNNs model disordered and difficult to extract specific semantic information. The “\textit{baseline+CL}” is slightly worse than the “\textit{baseline+CL+STL}”, indirectly proving the importance of optimizing SNN workloads.

\textbf{\textit{Time window size T.}} Theoretically, the larger the time window size, the better the performance. However, current SNNs have signal transmission defects. To this end, we made strategic workload adjustments to the model tasks and output methods. Fig. \ref{fig_3} visualizes the impact of different window sizes $T$ on model performance on the VQAv2 dataset. We find that the scores are close when the window size is 10 and 20, which is since the designed discrete semantic module only needs to consider the variables in memory.

\textbf{\textit{Semantic memory unit (SMU) evaluation.}} We set two states, “\textit{No SMU}” and “\textit{SMU}”, to evaluate the effectiveness of the semantic memory unit. As shown in Fig. \ref{fig_4}, the model with “\textit{SMU}” is better than the “\textit{No SMU}” state on four different VL datasets. Since there are differences in the spatial vector distributions between discrete semantic and continuous semantic encoders, we set the same initial values for gradient descent to enrich the feature mapping from SNN to the semantic memory unit.
%%%%%%%%%Table11%%%%%%%%
\begin{table*}[!t]
        \centering
	\caption{Performance evaluation of ITR task on the MSCOCO and Flickr30k dataset.}
	\label{table_11}
	\setlength{\tabcolsep}{3pt} %表格线加粗
	\renewcommand\arraystretch{1.5} %增加表格行距
	\begin{tabular}{|m{2.3cm}<{\centering}|m{1.2cm}<{\centering}|m{0.9cm}<{\centering}|m{0.9cm}<{\centering}|m{0.9cm}<{\centering}|m{0.9cm}<{\centering}|m{0.9cm}<{\centering}|m{0.9cm}<{\centering}|m{0.9cm}<{\centering}|m{0.9cm}<{\centering}|m{0.9cm}<{\centering}|m{0.9cm}<{\centering}|m{0.9cm}<{\centering}|m{0.9cm}<{\centering}|}
	%保持水平\垂直居中
		\hline 
		\multirow{3}{*}{Model} & \multirow{3}{*}{Parameter} &\multicolumn{6}{c|}{MSCOCO (5K test set)} &\multicolumn{6}{c|}{Flickr30k (1K test set)} \\ \cline{3-14}
		&  &\multicolumn{3}{c|}{IR} &\multicolumn{3}{c|}{TR} &\multicolumn{3}{c|}{IR} &\multicolumn{3}{c|}{TR} \\ \cline{3-14}
		&  & R@1 &R@5 &R@10 & R@1 &R@5 &R@10 &R@1 &R@5 &R@10 & R@1 &R@5 &R@10 \\ \hline
		RCAR \cite{ref62} &- &44.30 &73.20 & 83.20 &61.30 &86.10 &92.60 & 62.60 &85.80 & 91.10 & 82.30 & 96.00 & 98.40 \\ \hline
		PTP-ViLT \cite{ref46} &87M &45.30 &79.10 &88.40  &67.10 &90.50 &94.30 &68.80  &91.40 &95.30  &85.20  &96.90  &98.50  \\ \hline
		UNITER$_{large}$ \cite{ref5} &300M &52.90 &79.90 &88.00  &65.70 &88.60 &93.80 &75.60  &94.10 &96.80  &87.30  &98.00  &99.20  \\ \hline
		VISTA$_{large}$ \cite{ref63} &140M &52.60 &79.60 &87.60  &68.90 &90.10 &95.40 &75.80  &94.20 &96.90  &89.50  &98.40  &99.60  \\ \hline	
		METER-Swin$_{base}$ \cite{ref58} &380M &54.80 &81.40 &89.30  &72.90 &92.00 &96.20 &79.00  &95.50 &98.00  &92.40  &99.00  &99.50  \\ \hline
		ALBEF$_{4M}$	\cite{ref34} &210M &56.80 &81.50 &89.20  &73.10 &91.40 &96.00 &82.80  &96.70 &98.40  &94.30  &99.40  &99.80  \\ \hline	
		GSHN (Ours) &104.33M &65.11 &86.62 &91.76 &82.25 &95.89 &98.04 &79.10  &94.69 &98.09  &90.03  &98.42  &99.60  \\ \hline
\end{tabular}
\end{table*}

%%%%%%%%%Table12%%%%%%%%
\begin{table}[!t]
        \centering
	\caption{Performance evaluation of VE task on the SNLI-VE dataset.}
	\label{table_12}
	\setlength{\tabcolsep}{3pt} %表格线加粗
	\renewcommand\arraystretch{1.5} %增加表格行距
	\begin{tabular}{|m{2.5cm}<{\centering}|m{1.8cm}<{\centering}|m{1.7cm}<{\centering}|m{1.7cm}<{\centering}|}
	%保持水平\垂直居中
		\hline 
		Model &Parameter &test & val   \\ \hline
		ALBEF$_{14M}$  \cite{ref34}  &210M & 80.91 &80.80   \\ \hline
		METER-CLIP  \cite{ref58} &380M &81.19  &80.86  \\ \hline
		SimVLM$_{huge}$  \cite{ref59} &1.2B &86.32  &86.21   \\ \hline
		CoCa  \cite{ref60} &2.1B &87.10  &87.00  \\ \hline
		OFA$_{base}$  \cite{ref61} &182M & 89.20 &89.30  \\ \hline
		GSHN (Ours) &104.33M &89.22   &89.61 \\ \hline			
\end{tabular}
\end{table}

\textbf{\textit{Speed analysis.}} Considering the actual resource requirements, we modeled cases with five different strategies, namely “\textit{GAT}”, “\textit{SNN}”, “\textit{GAT+GAT}”, “\textit{SNN+SNN}” and “\textit{GSHN}”. From Table \ref{table_7}, we have some observations upon inspection. First, the computational efficiency of “\textit{SNN}” is slightly higher than that of “\textit{GAT}”, and there is no intuitively significant difference between them. It is worth noting that we choose STBP-tdBN to perform batch normalization of SNN to accelerate model convergence. This operation not only reduces the training epochs but also reduces the average epoch training time. “\textit{GSHN}” refers to the hybrid architecture of GAT and SNN, which can perform independent parameter operations in parallel. Furthermore, the running time of the model is reduced when it transitions from a single structure to a parallel structure, rather than simply performing a mathematical operation of multiplying by two. Second, the performance of “\textit{GAT}” and “\textit{GAT+GAT}” is close, because “\textit{GAT}” can process the continuous features of graph nodes, which is enough to capture the detailed features. In contrast, both “\textit{SNN}” and “\textit{SNN+SNN}” performed poorly due to the lack of detailed feature descriptions. “\textit{GSHN}” outperformed “\textit{GAT}” and “\textit{GAT+GAT}”, which indirectly proved the effectiveness and necessity of “\textit{SNN}”. This is due to the spatiotemporal properties of SNN and its ability to activate different pulses to model similar semantic representations.

\subsection{Downstream tasks and results}
This sub-section provides experimental results on four different VL tasks. The evaluation tests for the VQA task can be referred to Table \ref{table_8}, Table \ref{table_9}, and Table \ref{table_10}. Specifically, Table \ref{table_8} provides comparative experiments of different state-of-the-art methods on the VQAv2 dataset. They achieve the best performance by improving the model performance to 73.39\% and 73.56\% on the \textit{dev} and \textit{std} sets, respectively. We also evaluate the test on the VQA-CP v2 dataset to demonstrate the generality of GSHN. Table \ref{table_9} shows the performance comparison of different question types. Here, our model achieves an accuracy of 58.95\%, significantly outperforming other compared methods. In addition, on the GQA dataset tested in Table \ref{table_10}, GSHN also achieves the best performance (65.09\% and 64.45\%) compared to other methods.

%%%%%%%%%Table13%%%%%%%%
\begin{table}[!t]
        \centering
	\caption{Performance evaluation of VR task on the NLVR$^2$ dataset.}
	\label{table_13}
	\setlength{\tabcolsep}{3pt} %表格线加粗
	\renewcommand\arraystretch{1.5} %增加表格行距
	\begin{tabular}{|m{2.5cm}<{\centering}|m{1.8cm}<{\centering}|m{1.7cm}<{\centering}|m{1.7cm}<{\centering}|}
	%保持水平\垂直居中
		\hline 
		Model &Parameter &test-P & dev   \\ \hline
		METER-Swin \cite{ref58}  &380M & 82.47 &82.23   \\ \hline
		METER-CLIP \cite{ref58} &380M &83.05  &82.33  \\ \hline
		VinVL$_{large}$ \cite{ref57} &550M &83.98 &82.67 \\ \hline
		VK-OOD-s$_{ViLT}$ \cite{ref64} &87.4M &- &84.30 \\ \hline
		X-VLM$_{base}$  \cite{ref65} &216M &84.21  &84.16   \\ \hline
		SimVLM$_{large}$  \cite{ref59} &1.2B &84.84  &84.13   \\ \hline
		PTP-BLIP$_{14M}$  \cite{ref46} &220M & 83.17 &84.55  \\ \hline		
		GSHN (Ours) &104.33M &84.85   &84.57 \\ \hline			
\end{tabular}
\end{table}

For downstream ITR tasks, we evaluate the performance comparison with the state-of-the-art methods in Table \ref{table_11}. Our model outperforms the comparison methods in most indicators on MSCOCO and Flickr30k, which indirectly shows that GSHN can train better image-text representations to align semantics. Here, we use Recall @K as the evaluation metric. R@K (K=1,5,10) represents the percentage of correct matches in the top K list. The higher the R@K, the better the performance. The results show that GSHN still has certain advantages. Table \ref{table_12} tests the performance of the model in the VE task. Even compared with the huge-capacity SimVLM methods, our model still achieves good performance, with \textit{test} by 2.9\% (from 86.32\% to 89.22\%), and \textit{val} by 3.4\% (from 86.21\% to 89.61\%). This verifies that GSHN is beneficial for aligning VL embeddings in feature space and promotes improvements on more fine-grained visual reasoning tasks. 

From Table \ref{table_13}, we perform the Visual Reasoning task on the NLVR$^2$ dataset. It is worth noting that the batch size of NLVR$^2$ is half of that of VQA. GSHN achieves an accuracy of 84.85\% and 84.57\% on \textit{test-P} and \textit{dev}, respectively. The results are slightly better than those provided by PTP-BLIP \cite{ref46}, with an absolute gain of 1.68\% and 0.02\% on \textit{test-P} and \textit{dev}, respectively. The above comparative experiments fully demonstrate the advantages of end-to-end GSHN, which narrows the spatial differences in the VL alignment process and is beneficial to the healthy development of the model.

\section{Conclusion}
In this paper, we design a novel visual semantics module that captures fine-grained semantic features based on segmented object instances as visual tokenized embeddings. More interestingly, we construct a comprehensive and compact graph spike hybrid network (GSHN) in the module, which can simultaneously encode continuous and discrete semantic features to optimize VL-aligned representations. Due to the sparsity and energy efficiency of SNNs, GSHN can perform sparse attention coefficients to selectively aggregate neighbor node features, alleviating the impact of noise on the model. Furthermore, we have effectively verified it on public datasets of multiple VL downstream tasks and demonstrated multiple competitive results. Considering the computational consumption of the model, the efficiency of spike-based temporal encoding will be further improved in the future, and the model will be given stronger generalization capabilities.

\vfill
\end{document}